\documentclass{article}
\usepackage{amssymb}
\usepackage{adjustbox}
\usepackage{booktabs,graphicx,makecell,siunitx,eqparbox}
\usepackage{tabularx}
\usepackage{footnote}
\usepackage{amsmath,amssymb}
\usepackage{wrapfig}
\usepackage{balance}
\usepackage{subcaption}
\usepackage{hyperref}
\usepackage{pdfpages}
\usepackage[capitalize,noabbrev]{cleveref}

\usepackage[acronym,shortcuts]{glossaries}

\usepackage[final]{corl_2025} 

\title{FFHFlow: Diverse and Uncertainty-Aware Dexterous Grasp Generation via Flow Variational Inference}

\author{
Qian~Feng\thanks{Equal Contributions, author ordering decided via coin-tossing: \texttt{\{qian, jianxiang\}.feng@tum.de}. \dag work done while at Agile Robots SE.
$^{1}$Amigos Robots $^{2}$Agile Robots SE $^{3}$School of Information Computation and Technology, Technical University of Munich (TUM) $^{4}$Institute of Robotics and Mechatronics, German Aerospace Center (DLR) $^{5}$Department of Informatics, Karlsruhe Institute of Technology (KIT)}
$^{~\dag1,3}$,
Jianxiang~Feng$^{*}$$^{2,3}$,
Zhaopeng~Chen$^{2,}$,
Rudolph~Triebel$^{4,}$$^{5}$ and Alois Knoll$^{3}$
}

\begin{document}
\maketitle



\def\ie{\textit{i.e., }}
\def\eg{\textit{e.g., }}
\def\FirstFlowName{\textit{FFHFlow-cnf}~}
\def\SecondFlowName{\textit{FFHFlow-lvm}~}

\def\graspVar{\mathbf{g}}
\def\pcdVar{\mathbf{x}}
\def\latentVar{\mathbf{z}}
\def\GeneratorParam{\theta}
\def\InferenceParam{\phi}

\def\flowNum{n}
\def\latentDim{l}
\def\featDim{d}
\def\baseParam{\psi}
\def\baseVarCap{U}
\def\baseVar{\bb{u}}
\def\flowFunc{T}
\def\subflowFunc{t}
\def\flowParam{\phi}
\def\detFunc{F} 
\def\detParam{\theta}

\def\generatSet{\mathbf{G_{gen}}}
\def\generatGrasp{\mathbf{g_{gen}}}
\def\gtSet{\mathbf{G_{gt}}}

\newcommand{\rebuttal}[1]{\textcolor{black}{#1}}
\newcommand{\bb}[1]{\textbf{#1}}
\newcommand{\pb}[1]{\textbf{#1}^\prime}
\newcommand{\mb}[1]{\mathcal{#1}}
\newcommand{\todo}[1]{\textbf{\color{red}{TODO: \textit{#1}}}}
\newcommand{\ra}[1]{\renewcommand{\arraystretch}{#1}}

\newcommand\myeq{\mkern1.5mu{=}\mkern1.5mu}


\begin{abstract}
Synthesizing diverse, uncertainty-aware grasps for multi-fingered hands from partial observations remains a critical challenge in robot learning. 
Prior generative methods struggle to model the intricate grasp distribution of dexterous hands and often fail to reason about shape uncertainty inherent in partial point clouds, leading to unreliable or overly conservative grasps. 
We propose FFHFlow, a flow-based variational framework that generates diverse multi-finger grasps while explicitly quantifying perceptual uncertainty in the partial point clouds. 
Our approach leverages a normalizing flow-based deep latent variable model to learn a hierarchical grasp manifold, overcoming the mode collapse and rigid prior limitations of conditional Variational Autoencoders (cVAEs). 
By exploiting the invertibility and exact likelihoods of flows, FFHFlow introspects shape uncertainty in partial observations and identifies novel object structures, enabling risk-aware grasp synthesis. 
To further enhance reliability, we integrate a discriminative grasp evaluator with the flow likelihoods, formulating an uncertainty-aware ranking strategy that prioritizes grasps robust to shape ambiguity. Extensive experiments in simulation and real-world setups demonstrate that FFHFlow outperforms state-of-the-art baselines (including diffusion models) in grasp diversity and success rate, while achieving run-time efficient sampling. 
We also showcase its practical value in cluttered and confined environments, where diversity-driven sampling excels by mitigating collisions. 
(Project Page: \href{URL}{https://sites.google.com/view/ffhflow/home/}) 
\end{abstract}

\keywords{Dexterous Grasping, Normalizing Flows, Uncertainty-Awareness}

\newglossaryentry{auc}{name=AUC, description={Area Under Curve},first={Area Under Curve (AUC)}}
\newglossaryentry{auroc}{name=AUROC, description={Area Under Receiver Operation Curve},first={Area Under Receiver Operation Curve (AUROC)}}
\newglossaryentry{aevb}{name=AEVB, description={Auto-encoding Variational Bayes},first={Auto-encoding Variational Bayes (AEVB))}}

\newglossaryentry{bnn}{name=BNNs, description={Bayesian Neural Networks},first={Bayesian Neural Networks (BNNs)}}
\newglossaryentry{bps}{name=BPS, description={Basis Point Set},first={Basis Point Set (BPS)}}
\newglossaryentry{cgm}{name=CGMs, description={Conditional Generative Models},first={Conditional Generative Models (CGMs)}}
\newglossaryentry{cvae}{name=cVAE, description={Conditional Variational Autoencoder},first={Conditional Variational Autoencoder(cVAE)}}
\newglossaryentry{ci}{name=CI, description={Mutual Cross-Information},first={Mutual Cross-Information (CI)}}
\newglossaryentry{cov}{name=Cov, description={Coverage},first={Coverage (Cov)}}
\newglossaryentry{dof}{name=DoF, description={Degrees of Freedom},first={Degrees of Freedom (DoF)}}
\newglossaryentry{dlvm}{name=DLVMs, description={Deep Latent Variable Models},first={Deep Latent Variable Models (DLVMs)}}
\newglossaryentry{dl}{name=DL, description={Deep Learning},first={Deep Learning (DL)}}
\newglossaryentry{dnn}{name=DNNs, description={Deep Neural Networks},first={Deep Neural Networks (DNNs)}}
\newglossaryentry{dgm}{name=DGM, description={Deep Generative Models},first={Deep Generative Models (DGMs)}}
\newglossaryentry{ee}{name=EE, description={End Effectors},first={End Effectors (EE)}}
\newglossaryentry{fpr}{name=FPR, description={False Positive Rate},first={False Positive Rate (FPR)}}
\newglossaryentry{gan}{name=GANs, description={Generative Adversarial Networks},first={Generative Adversarial Networks (GANs)}}
\newglossaryentry{diffusion}{name=Diffusion, description={Diffusion},first={Diffusion}}
\newglossaryentry{gm}{name=MoG, description={Mixture of Gaussians Distributions},first={Mixture of Gaussians (MoG)}}
\newglossaryentry{gmm}{name=GMMs, description={Gaussian Mixture Models},first={Gaussian Mixture Models (GMMs)}}
\newglossaryentry{iid}{name=\textit{iid}, description={independent and identically distributed},first={independent and identically distributed (\textit{iid})}}
\newglossaryentry{ib}{name=IB, description={Information Bottleneck},first={Information Bottleneck (IB)}}
\newglossaryentry{id}{name=ID, description={In-Distribution},first={In-Distribution (ID)}}
\newglossaryentry{kld}{name=KLD, description={Kullback-Leibler Divergence},first={Kullback-Leibler Divergence (KLD)}}
\newglossaryentry{ll}{name=LL, description={Log-Likelihood},first={Log-Likelihood (LL)}}
\newglossaryentry{lars}{name=LARS, description={Learned accept/reject sampling},first={Learned accept/reject sampling (LARS)}}
\newglossaryentry{mle}{name=MLE, description={Maximum Likelihood Estimation},first={Maximum Likelihood Estimation (MLE)}}
\newglossaryentry{mi}{name=MI, description={Mutual Information},first={Mutual Information (MI)}}
\newglossaryentry{ml}{name=ML, description={Machine Learning},first={Machine Learning (ML)}}
\newglossaryentry{magd}{name=MAGD, description={Mean Absolute Grasp Deviation},first={Mean Absolute Grasp Deviation (MAGD)}}
\newglossaryentry{mlp}{name=MLP, description={Multi-Layer Perceptron},first={Multi-Layer Perceptron (MLP)}}
\newglossaryentry{nf}{name=NFs, description={Normalizing Flows},first={Normalizing Flows (NFs)}}
\newglossaryentry{cnf}{name=cNF, description={Conditional Normalizing Flow},first={Conditional Normalizing Flows (cNF)}}
\newglossaryentry{ood}{name=OOD, description={Out-of-Distribution},first={Out-of-Distribution (OOD)}}
\newglossaryentry{rsb}{name=RSB, description={Resampled Base Distributions},first={Resampled Base Distributions (RSB)}}
\newglossaryentry{sgvb}{name=SGVB, description={Stochastic Gradient Variational Bayes},first={Stochastic Gradient Variational Bayes (SGVB) }}
\newglossaryentry{tpr}{name=TPR, description={True Positive Rate},first={True Positive Rate(TPR)}}
\newglossaryentry{vae}{name=VAE, description={Variational Autoencoder},first={Variational Autoencoder (VAE)}}
\newglossaryentry{vi}{name=VI, description={Variational Inference},first={Variational Inference (VI)}}

\section{Introduction}
\vspace{-5pt}

Performing diverse dexterous grasps on various objects is important to realize advanced human-like robotic manipulation. 
However, achieving such capability remains challenging due to the high dimensionality of the hand configuration space and the high variability in the \textit{object shape}, not to mention the high perceptual uncertainty in the \textit{incomplete shape} from the partial observation.

Previous efforts to address this problem have garnered significant interest in \ac*{cvae}~\cite{mousavian20196, wei2022dvgg}.
However, its performance is hampered by the issues of \textit{mode collapse} ~\cite{mattei2018leveraging,zhao2019infovae,richardson2018gans} and the often-used \textit{overly simple prior}~\cite{casale2018gaussian,tomczak2018vae}.
These severely limit its ability to model the highly complex and inherently multi-modal grasp distribution. 
The lack of diversity significantly restricts the manipulability of robots operating in the common cluttered or confined space (\eg a two-tier shelf in \cref{fig:YCB_objects}) in daily life.
Recent advances in the application of diffusion models~\cite{weng2024dexdiffuser} can potentially mitigate this issue but are struggling to achieve satisfactory run-time efficiency.
More severely, only few prior studies~\cite{lundell2019robust, humt2023shape} considered the necessity of having \textit{introspective capabilities} against the perceptual uncertainty raised by partial observation. Unfortunately, such solutions come with a often slow shape completion module, 
\begin{wrapfigure}{r}{0.5\textwidth}
    \begin{center}
    \includegraphics[width=0.5\textwidth]{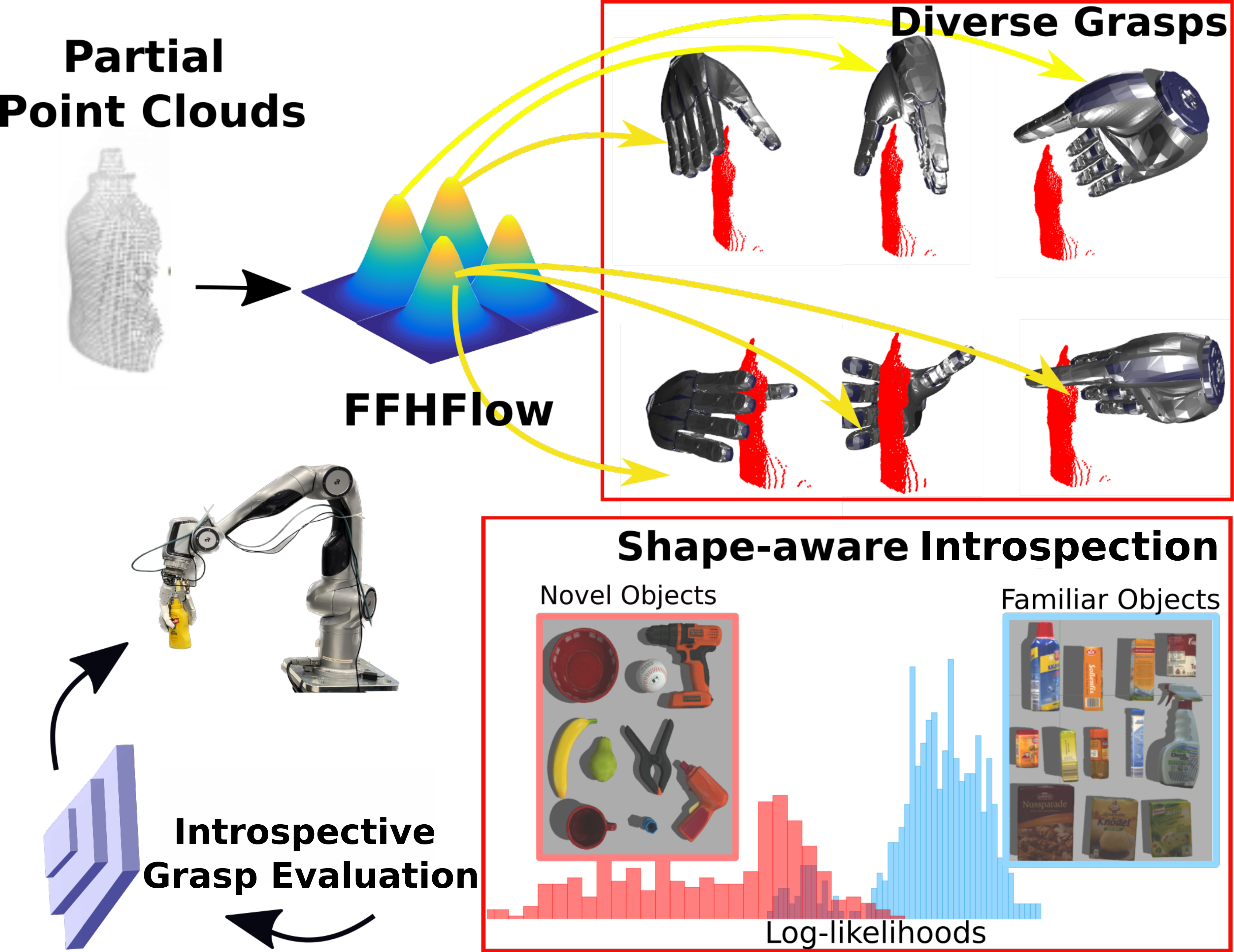}
    \end{center}
    \caption{\textbf{Method overview}: in this work, we propose a variational grasp sampler based on \ac*{nf}, generating \textit{diverse} dexterous grasps with shape-aware \textit{introspection}.
    }
    \label{fig:teaser_fig}
    \vspace{-15pt}
\end{wrapfigure}
not to mention the issue of Out-of-Distribution (OOD) objects with unknown object shapes uncovered by the training data, which easily cause unexpected consequences, \eg grasp failure.  

In this work, we introduce a novel flow-based variational grasp generation model to overcome the aforementioned challenges (\cref{fig:teaser_fig}).
To this end, we first observed deficient generalization gain by simply using \ac*{cnf} to model the point cloud-conditioned grasp distribution due to suboptimal latent features. 
To address this, we devise \SecondFlowName based on a novel flow-based \ac*{dlvm} (\cref{fig:method_fig}).
In this model, we first replace the commonly used isotropic Gaussian with an expressive \textit{input-dependent} NF prior~\cite{zhao2019infovae,tomczak2018vae} (Prior Flow). 
Further, we construct an elastic non-Gaussian likelihood function~\cite{mattei2018leveraging} (Grasp Flow) with another NF within the framework of \ac*{dlvm}.
With this design, we aim to conquer the over-regularization induced by the simple \textit{input-independent} prior and the restricted form of the likelihood function (\eg isotropic Gaussian) in \ac*{cvae}.  

To facilitate shape-aware introspection, inspired by~\cite{feng2023topology}, we exploit the exact likelihoods of \ac*{nf} to represent the perceptual uncertainty (lower the more uncertain) related to the generated grasps. 
Specifically, the likelihoods of Grasp Flow account for the \textit{incomplete shape} due to partial observability, namely \textbf{view uncertainty} (illustrated in \cref{fig:grasp_flow}). It predicts lower values for grasps toward the invisible views. 
Meanwhile, Prior Flow assigns lower likelihoods for OOD objects with \textit{distinct shapes} to the objects in the training data, representing the \textbf{object uncertainty}. 
More noteworthy, the object uncertainty can be applied to promote robustness by detecting objects with novel geometry, which can easily result in grasp failures.
Moreover, we leverage the view uncertainty to assist grasp evaluation and propose a new evaluation strategy based on a discriminative grasp evaluator.

To summarize, we contribute with \textbf{(1)} a novel flow-based \ac*{dlvm} that can address the limitations of its alternatives, such as \ac*{cvae} and diffusion models,
\textbf{(2)} a new way to represent the perceptual uncertainties based on the flow likelihoods, leading to an uncertainty-aware grasp evaluation strategy and
\textbf{(3)} a comprehensive experimental study both in simulation and on the real robot to verify the proposed idea, \ie object grasping from free space to clutter and confined space. 

\section{Related Work}
\vspace{-10pt}
\label{sec:related_work}
\textbf{Learning-based Grasp Synthesis.} 
Deep generative models such as \ac*{cvae}s~\cite{mousavian20196, ffhnet, wei2022dvgg, zhang2024ffhclutter}, auto-regressive models~\cite{9981133}, \ac*{cnf}s~\cite{yan2019learning, xu2023unidexgrasp, lim2024equigraspflow}, diffusion models~\cite{weng2024dexdiffuser, zhang2024dexgraspnet20,carvalho2024graspdiffusion} and \ac*{gan}~\cite{patzelt2019conditional, feng2024dexgangrasp} are widely adopted in grasp synthesis with two-jaw gripper and multi-fingered hands. 
Among them, some~\cite{wei2022dvgg, 9981133} address the partial observation challenge by employing time-intensive shape completion modules, which significantly slow down inference, reducing their practical applicability. Other approaches either assume the availability of complete object point clouds~\cite{xu2023unidexgrasp}, a limitation in real-world scenarios, or only focus on a 2-jaw gripper~\cite{lim2024equigraspflow}.
Similarly, diffusion models~\cite{weng2024dexdiffuser, zhang2024dexgraspnet20,carvalho2024graspdiffusion} face runtime inefficiency due to their iterative denoising processes, even though there are some techniques~\cite{rombach2022latentdiffusion,song2022ddim} to accelerate it.
In contrast, our method, similar to~\cite{yan2019learning, ffhnet}, eliminates the need for resource-intensive shape completion models by encoding partial observation information directly into latent variables. This lightweight approach allows our model to effectively handle partial observations while maintaining efficiency and diversity in grasp synthesis, enabling real-world applications such as grasping in clutter and dynamic grasping~\cite{burkhardt2024multifingered}. 

\textbf{Uncertainty-Aware Grasping.} 
Prior research on uncertainty-aware grasping with deep learning (DL) is primarily focused on suction cups~\cite{SuctionUncExplore2024, SuctionGraspUnc2024} and parallel-jaw grippers~\cite{9341056, stephan2022importance, shi2024vmf}.
Their motivations and applications stem from the characteristic of uncertainty estimation in DL~\cite{gawlikowski2023survey}, ranging from efficient adaptation~\cite{9341056} and exploration~\cite{SuctionUncExplore2024} to  grasp evaluation~\cite{stephan2022importance, SuctionGraspUnc2024, shi2024vmf, lum2024getagrip,zheng2024evaluating}.
Although with encouraging results, their investigation is limited to a rather simplistic setting, i.e., either using simplified end-effectors or restricting the analysis to 2D spaces.
In contrast, non-DL based uncertainty-aware grasping considers shape analysis~\cite{shapeUnc2007,gaussianShape2011,chen2018probabilistic} and multi-fingered hands~\cite{li2016dexterous, chen2024springgrasp}, which exhibits an understudied gap.
Our work aims to bridge this gap by developing a holistic learning-based grasping model that can retrospectively reason about shape-aware uncertainty against partially observed and unknown objects.
Some concurrent works~\cite{lundell2019robust, humt2023shape} share a similar spirit but rely on a time-consuming shape completion module, while our work is more real-time capable.

\textbf{Normalizing Flows.}
Unlike other \ac*{dgm}, flow-based models~\cite{dinh2016density, kingma2018glow} can perform both exact likelihood evaluation and efficient sampling simultaneously.
More noteworthy, \ac*{nf} can be trained more stably when compared with \ac*{gan}~\cite{mescheder2018training}, perform better against the notorious \textit{mode collapse} problem in both \ac*{gan} and \ac*{vae}~\cite{richardson2018gans, mattei2018leveraging, zhao2019infovae}. 
Moreover, we focus on discrete \ac*{nf}, which is more run-time efficient than its continuous version based on flow matching~\cite{lipman2023flowmatching} and does not require long trajectories in the de-nosing process like diffusion models. 
These appealing properties render \ac*{nf} a promising tool for fast and effective probabilistic inference~\cite{papamakarios2021normalizing}.
In addition, \ac*{nf} found successful applications in point clouds processing~\cite{klokov2020discrete, postels2021go}, 
feasibility learning~\cite{feng2023density}, 
uncertainty estimation~\cite{charpentier2020posterior, postels2020hidden}, Out-of-distribution detection~\cite{kirichenko2020normalizing, feng2023topology}.
Inspired by the probabilistic nature of \ac*{nf}, we attempt to leverage this model to capture the multi-modal and complex grasp distribution and establish shape-aware introspective capability.


\section{Preliminaries}
\vspace{-5pt}

\textbf{Deep Latent Variable Models (DLVMs).}
\label{sec:intro_dlvm}
In the context of modeling the unknown true data distribution $p^*(\pcdVar)$ with a model $p_{\GeneratorParam}(\pcdVar)$ parameterized by $\GeneratorParam$ based on a dataset $\mathcal{D}=\{\pcdVar_{i}\}^{N}_{i=1}$, latent variables $\latentVar$ are usually introduced for discovering fine-grained factors controlling the data generating process or increasing the expressivity of the model $p_{\GeneratorParam}(\pcdVar)$.
Latent variables $\{\latentVar_{i}\}^{N}_{i=1}$ are part of the model but hidden and unobservable in the dataset.  
The resulting marginal probability is: $p_{\GeneratorParam}({\pcdVar}) = \int p_{\GeneratorParam}(\pcdVar|\latentVar)p_{\GeneratorParam}(\latentVar)d\latentVar$.
When $p_{\GeneratorParam}(\pcdVar, \latentVar)$ is parameterized by \ac*{dnn}, we term the model \ac*{dlvm}~\cite{kingma2019introduction, mattei2018leveraging}.
The difficulty of learning such models with \ac*{mle} lies in the intractability of the integral in the marginal probability for not having an analytic solution or efficient estimator.
To remedy this, \ac*{vi}~\cite{blei2017variational} provides a tractable lower bound of the marginal likelihood $p_{\GeneratorParam}(\pcdVar)$ to optimize by approximating the real posterior $p_{\GeneratorParam}(\latentVar|\pcdVar)$ with an approximate one $q_{\InferenceParam}(\latentVar|\pcdVar)$:
\begin{equation}
    \label{eq:eblo}
    \log{p_{\GeneratorParam}(\pcdVar)} \ge \mathbb{E}_{q_{\InferenceParam}(\latentVar|\pcdVar)}[\log{p_{\GeneratorParam}(\pcdVar|\latentVar)}+\log{\frac{p_{\GeneratorParam}(\latentVar)}{q_{\InferenceParam}(\latentVar|\pcdVar)}}]. 
\end{equation}
When $q_{\InferenceParam}(\latentVar|\pcdVar)$ and ${p_{\GeneratorParam}(\pcdVar|\latentVar)}$, are approximated by \ac*{dnn} with an isotropic Gaussian as prior $p_{\GeneratorParam}(\latentVar)$, we obtain the well-known instance of \ac*{dlvm}, the \ac*{vae} model~\cite{kingma2019introduction}.

\begin{figure*}[t!]
    \vspace{6pt}
    \centering
    \includegraphics[width=0.95\textwidth]{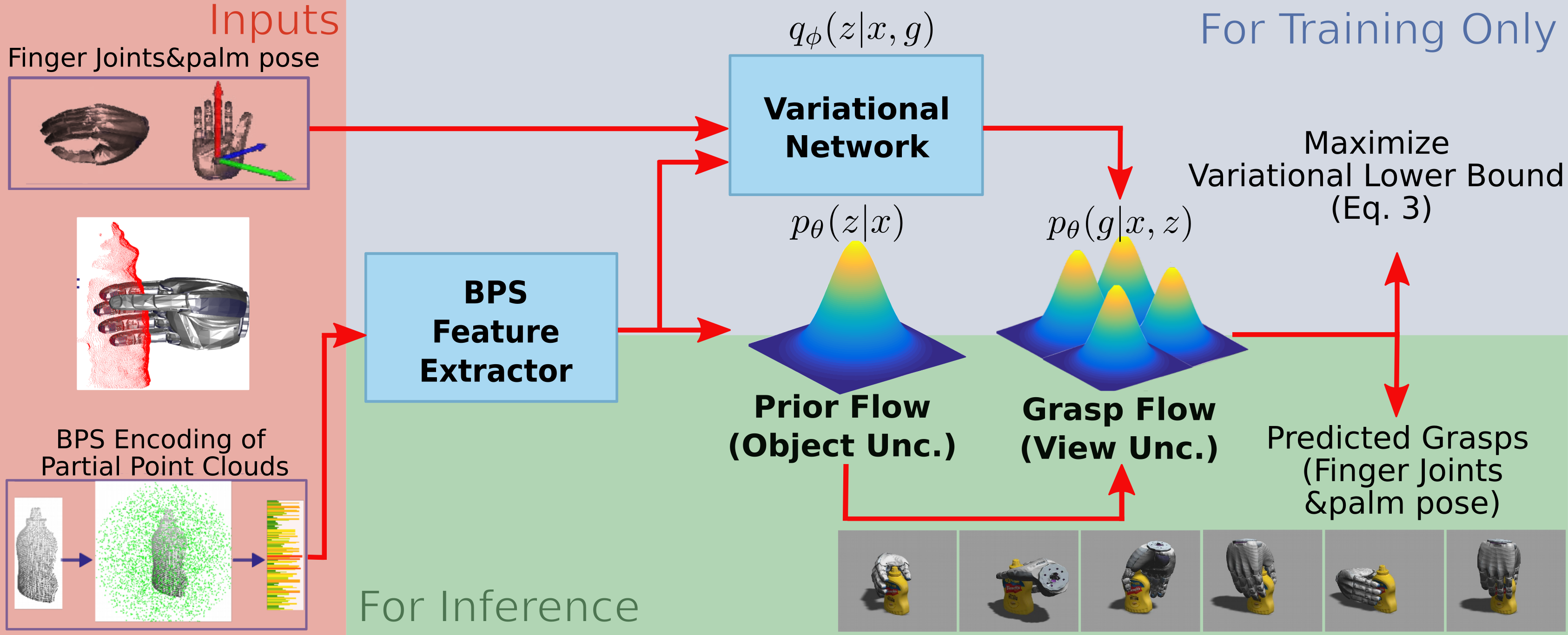}
    \caption{\textbf{Model Architecture}:
    at training time, the approximate posterior of the latent variable is inferred while the derived variational lower bound is maximized. 
    During inference, Prior flow generates latent samples conditioned on the input point clouds, based on which Grasp Flow predicts the final grasps 
    (best viewed in color).}
    \label{fig:method_fig}
    \vspace{-15pt}
\end{figure*}
\textbf{Normalizing Flows.}
\label{sec:intro_nf}
\ac*{nf} are known to be universal distribution approximators~\cite{papamakarios2021normalizing}.
That is, they can model any complex target distribution $p^{*}(\pcdVar)$ on a space $\mb{R}^\featDim$ by defining $\pcdVar$ as a transformation $\flowFunc_\GeneratorParam: \mb{R}^\featDim \rightarrow \mb{R}^\featDim$ parameterized by $\GeneratorParam$ from a well-defined base distribution $p_{u}(\baseVar)$: $\pcdVar = \flowFunc_\GeneratorParam(\baseVar) \text{~~where~~} \baseVar \sim p_u(\baseVar)\,,$
where $\baseVar \in \mb{R}^\featDim$ and $p_u$ is commonly chosen as a unit Gaussian.
By designing $T_{\GeneratorParam}$ to be a \textit{diffeomorphism}, that is, a bijection where both $T_{\GeneratorParam}$ and $T^{-1}_{\GeneratorParam}$ are differentiable, we can compute the likelihood of the input $\pcdVar$ exactly based on the change-of-variables formula~\cite{bogachev2007measure}: $ p_{\GeneratorParam}(\pcdVar) =  p_u(\flowFunc_{\GeneratorParam}^{-1}(\pcdVar))|\det(J_{T_{\GeneratorParam}^{-1}}(\pcdVar))|$,
where $J_{T_{\GeneratorParam}^{-1}}(\pcdVar) \in \mb{R}^{\featDim \times \featDim}$ is the Jacobian of the inverse $\flowFunc_{\GeneratorParam}^{-1}$ with respect to $\pcdVar$.
The transformation $\flowFunc_\GeneratorParam$ can be constructed by composing a series of bijective maps denoted by $\subflowFunc_{i}$,
$\flowFunc_\GeneratorParam = \subflowFunc_1 \circ \subflowFunc_2 \circ ...  \circ \subflowFunc_\flowNum$.
When the target distribution is unknown, but samples thereof are available, we can estimate $\GeneratorParam$ by minimizing the forward \ac*{kld}, equivalent to minimizing the negative expected \ac*{ll}:
\def\Jdet{|\det(J_{T_{\GeneratorParam}^{-1}}(\pcdVar))|}
\def\baseL{(p_{u}(\flowFunc_{\GeneratorParam}^{-1}(\pcdVar)))}
\begin{equation}
    \label{eq:nf_mle}
    \operatorname*{argmin}_{\GeneratorParam}\left[-\mathbb{E}_{p^*(\pcdVar)}[ \log \baseL + \log \Jdet ]\right].
\end{equation}

\vspace{-20pt}
\section{Flow-based Grasp Synthesis}
\vspace{-5pt}

\subsection{Problem Formulation}
\vspace{-5pt}
This work aims to generate diverse grasps given a \textit{partial} point cloud denoted by $\pcdVar \in \mathbb{R}^{N\times3}$.
A grasp configuration $\mathbf{g} \in \mathbb{R}^d$ is represented by the 15-DOF hand joint configuration $\boldsymbol{j} \in \mathbb{R}^{15}$ and the 6D palm pose $(\mathbf{R},\mathbf{t}) \in SE(3)$.
Given an empirical dataset of $N$ objects with $N_i$ corresponding possible grasps $\mathcal{D} = \{\pcdVar_i, \{\graspVar_{ik}\}_{k=1}^{N_i}\}_{i=1}^N$ drawn from the ground truth conditional distribution $p^*(\graspVar|\pcdVar)$, we train a probabilistic model $p_{\GeneratorParam}(\graspVar|\pcdVar)$ parameterized by $\GeneratorParam$ to approximate $p^*(\graspVar|\pcdVar)$.
To facilitate shape-aware introspection~\cite{feng2019}, we exploit the log-likelihoods of $p_{\GeneratorParam}(\graspVar|\pcdVar)$ to represent the uncertainty caused by incomplete point clouds and unknown object shapes. 

\vspace{-5pt}
\subsection{Flow-based Grasp Sampler: FFHFlow-cnf}
\begin{wrapfigure}{r}{0.6\textwidth}
    \centering
    \begin{subfigure}{0.15\textwidth}
        \includegraphics[width=2cm, height=2.8cm]{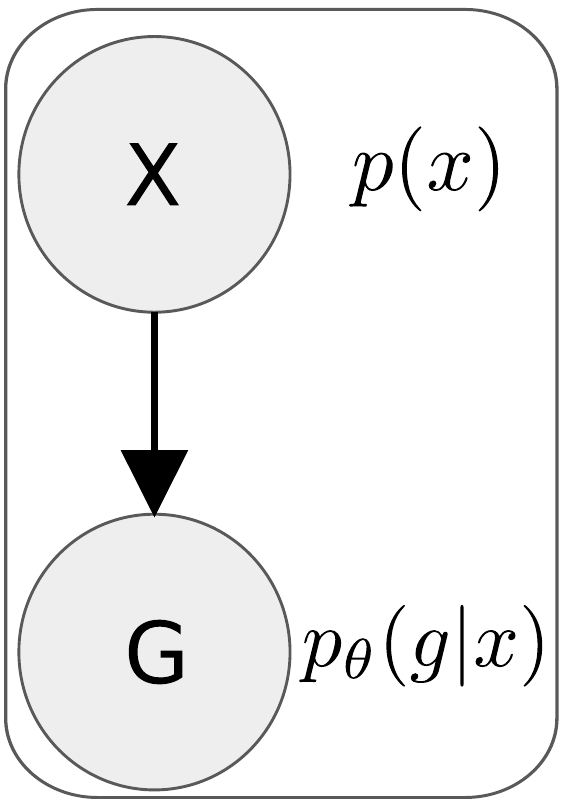}
        \caption{cNF}
        \label{fig:graphical_models:a}
    \end{subfigure}
    \hfill
    \begin{subfigure}{0.2\textwidth}
        \includegraphics[width=3cm, height=2.8cm]{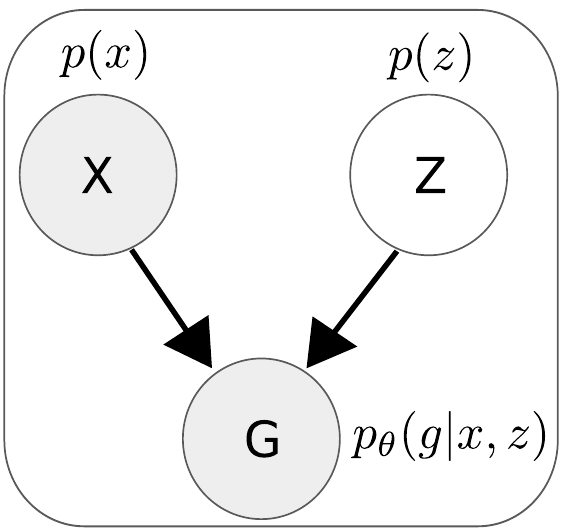}
        \caption{cVAE}
        \label{fig:graphical_models:b}
    \end{subfigure}
    \hfill
    \begin{subfigure}{0.2\textwidth}
        \includegraphics[width=3cm, height=2.8cm]{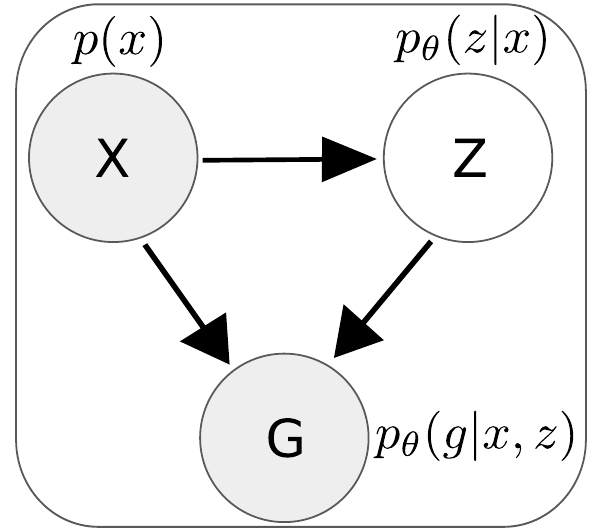}
        \caption{FFHFlow-lvm}
        \label{fig:graphical_models:c}
    \end{subfigure}
    \caption{\textbf{Graphical illustration}.
    Shaded and unshaded circles denote observed and hidden variables, while arrows for dependencies. 
    When compared to (a), \ac*{dlvm} (b) and (c) have an extra hidden variable $\latentVar$ for expressive latents learning. 
    Though, the prior of $\latentVar$ in cVAE (b) is an \textit{input-independent} and simplistic Gaussian, while that of ours (c) is an \textit{input-dependent} and more elastic cNF.}
    \vspace{-15pt}
    \label{fig:graphical_models}
\end{wrapfigure}

\vspace{-5pt}
\label{sec:cond_nf}
A straightforward idea to learn the conditional distribution $p_{\GeneratorParam}(\graspVar|\pcdVar)$ is directly employing the \ac*{cnf}~\cite{winkler2019learning} without defining hidden variables on the latent space shown in \cref{fig:graphical_models:a}.
To this end, we condition the flow transformation $\flowFunc_\GeneratorParam$ and the base distribution $p_u$ on the object point clouds $\pcdVar$, namely $\flowFunc_{\GeneratorParam|\pcdVar}:\mb{R}^\featDim \times \mb{R}^\latentDim \rightarrow \mb{R}^\featDim$ 
, where $\latentDim$ is the dimensionality of point cloud features and $\featDim$ for the grasp representation.
We encode each point cloud with a fixed \ac*{bps} according to \cite{prokudin2019efficient}, resulting in a feature vector $\mathbf{x_b}\in\mathbb{R}^{s}$ of fixed length $s$, before being fed into the feature extractor network $f_{\InferenceParam}(\mathbf{x_b}): \mathbb{R}^{s} \rightarrow  \mathbb{R}^{\latentDim} $.
\label{sec:cnf_limitation}
\textbf{Limitation:}
Though \FirstFlowName achieved encouraging improvements in terms of diversity and accuracy when compared to the \ac*{cvae}-based approach, we found it less generalizable with limited performance gain. 
We attribute the problem to the inadequate expressivity of the latent feature (explained in Section 3.3 of the appendix), especially when the model needs to understand the complicated relationships between the grasps and the partially observed point clouds of different objects.
To address this problem, we introduce \SecondFlowName in the next sub-section, a flow-based variational sampler with a more expressive probabilistic representation in the latent space.

\vspace{-5pt}
\subsection{Flow-based Variational Grasp Sampler: FFHFlow-lvm}
\vspace{-5pt}
\label{sec:flow_variational}
Inspired by the success of leveraging \ac*{dlvm} for point cloud processing~\cite{klokov2020discrete} and grasp generation~\cite{mousavian20196,ffhnet, wei2022dvgg}, we devise a flow-based variational model that can induce expressive latent distribution for precise and diverse grasp generation.
Specifically, we seek to overcome the over-regularization by the simplistic prior and the latent feature collapse by the Gaussian observation model in \ac*{cvae}-based approaches~\cite{mousavian20196,ffhnet, wei2022dvgg}.
This is achieved by introducing an input-dependent and expressive prior and a flexible observation model based on a \ac*{cnf}, which can be optimized efficiently under the framework of \ac*{sgvb}~\cite{kingma2013auto}.

\subsubsection{Learning Grasp Distribution via DLVMs}
\vspace{-10pt}
Our main idea is to introduce a latent variable into \textit{FFHFlow-cnf} to increase the expressivity of the latent space.
With the latent variable $\latentVar$ shown in \cref{fig:graphical_models:c}, we have the following conditional likelihoods of the grasps $\graspVar$ given a partially observed point cloud $\pcdVar$. It can be factorized: $p_{\GeneratorParam}(\graspVar|\pcdVar) = \int p_{\GeneratorParam}(\graspVar|\pcdVar,\latentVar)p_{\GeneratorParam}(\latentVar|\pcdVar)d\latentVar$.
The likelihood is intractable due to the integral over the latent variables.
We first derive a tractable lower bound for optimization (derivation in the appendix).
Then, each component in the model is explained as follows how the pitfalls in \ac*{cvae} are overcome.
To note that $\GeneratorParam$ denote the parameters for both flows and $\InferenceParam$ for the variational network.

\textbf{Variational Lower Bound:}
With Jensen Inequality, we have the following variational lower bound with an approximate posterior of the latent variable $q_{\InferenceParam}(\latentVar|\pcdVar, \graspVar)$ (more details in the appendix):
\begin{align}
    \begin{split}
    \label{eq:nf_lvm_eblo}
    \log{p_{\GeneratorParam}(\graspVar|\pcdVar)} \ge  &\mathbb{E}_{q_{\InferenceParam}(\latentVar|\pcdVar, \graspVar)}[\log{p_{\GeneratorParam}(\graspVar|\pcdVar, \latentVar)}] - \beta KL(q_{\InferenceParam}(\latentVar|\pcdVar,\graspVar)||p_{\GeneratorParam}(\latentVar|\pcdVar)).
    \end{split}
\end{align}
where $\beta$ is a hyperparameter to control the extend of shape-aware information in the latents. 
In practice, the \ac*{kld} term is implemented as the sum of the negative entropy of the approximate posterior and the cross-entropy between the prior and the approximate posterior for convenience.
The effect of minimizing this term is to match the input-dependent prior $p_{\GeneratorParam}(\latentVar|\pcdVar)$ to the approximate posterior $q_{\InferenceParam}(\latentVar|\pcdVar,\graspVar)$, and meanwhile encourage the approximate posterior to be more diverse. 
A flexible prior is of vital importance at inference time as we resort to the Prior Flow for drawing latent samples.
These samples will be conditioned on the Grasp Flow for grasp generation (\cref{fig:method_fig}). 
Therefore, \textit{an input-dependent and learnable prior} is not only useful for increasing expressivity but also necessary for generating diverse grasps given an incomplete point cloud.

\textbf{Grasp Flow $p_{\GeneratorParam}(\graspVar|\pcdVar, \latentVar)$:}
\label{sec:cnf_lvm_generator}
Both \ac*{vae} and \ac*{gan} are reported to suffer from the \textit{mode collapse} problem~\cite{richardson2018gans}, which is also named \textit{Information Preference Property}~\cite{zhao2019infovae}.
It illustrates the inability of \ac*{vae} to capture the entire data distribution, \eg the complex multi-modal grasp distribution.
Whenever this happens, the latent variable $\latentVar$ is neglected by the powerful decoder and hence is \textit{uninformative} in terms of the data, \ie grasps and object point clouds.
It has been proved that the unbounded likelihood function is the crux of this problem, \eg an isotropic Gaussian~\cite{mattei2018leveraging}.
To mitigate this, we propose to learn an \ac*{cnf} for the likelihood function $p_{\GeneratorParam}(\graspVar|\pcdVar, \latentVar)$ by abuse of notations in \cref{sec:cond_nf}: $    p_{\GeneratorParam}(\graspVar|\pcdVar, \latentVar) = p_{u|\latentVar}(\flowFunc_{\GeneratorParam|\latentVar}^{-1}(\graspVar; \latentVar); \latentVar) |\det(J_{T_{\GeneratorParam|\latentVar}^{-1}}(\graspVar; \latentVar))|$,
where the base distribution $p_{u|\latentVar}:\mb{R}^\featDim \times \mb{R}^\latentDim \rightarrow \mb{R}$ is conditional on $\latentVar$.
For conciseness, $\pcdVar$ is not shown on the RHS as $\latentVar$ already subsumes the information from $\pcdVar$, so Grasp Flow is simplified as  $p_{\GeneratorParam}(\graspVar|\latentVar)$.
In contrast to \ac*{cvae}, there is no underlying assumption for the distribution form in \ac*{nf}, which allows the model to learn a more general likelihood function instead of an isotropic Gaussian. 
Meanwhile, the architecture of \ac*{nf} is restricted to be a diffeomorphism, which we anticipate to help alleviate the unboundness issue.

\textbf{Prior Flow $p_{\GeneratorParam}(\latentVar|\pcdVar)$:}
\label{sec: latent_flow_prior}
The overly-simple prior can induce excessive regularization, limiting the quality of the latent representation~\cite{casale2018gaussian, tomczak2018vae}, which can be observed in \ac*{cvae}-based approaches~\cite{mousavian20196,ffhnet, wei2022dvgg} with an \textit{input-independent} isotropic Gaussian as the prior (\cref{fig:graphical_models:b}). 
On the other hand, the assumption of input-independence for the prior~\cite{zhao2019infovae} poses difficulty in learning informative latent features.
To address this, we propose to utilize a second \ac*{cnf} for an \textit{input-dependent} prior distribution, in our case, a point cloud-dependent prior: $ p_{\GeneratorParam}(\latentVar|\pcdVar) = p_{u}(\flowFunc_{\GeneratorParam|\pcdVar}^{-1}(\latentVar; \pcdVar)) |\det(J_{T_{\GeneratorParam|\pcdVar}^{-1}}(\latentVar; \pcdVar))|$.

\textbf{Variational Network $q_{\InferenceParam}(\latentVar|\pcdVar,\graspVar)$:}
The variational network is designed to approximate the real but intractable posterior distribution $p_{\GeneratorParam}(\latentVar|\pcdVar, \graspVar)$ for amortized variational inference.
$p_\theta(\latentVar|\pcdVar, \graspVar)$ is defined within the DLVM according to the Bayes formula: $p_\theta(\latentVar|\pcdVar, \graspVar) = \frac{p_\theta(\graspVar|\latentVar,\pcdVar)p_\theta(\latentVar|\pcdVar)}{p_\theta(\graspVar|\pcdVar)}$, where ${p_\theta(\graspVar|\pcdVar)} = \int {p_\theta(\graspVar|\latentVar,\pcdVar)p_\theta(\latentVar|\pcdVar) d\latentVar}$ is the model evidence.
From a pragmatic perspective, this network should be a powerful feature extractor for the grasps and object point clouds. 
As it is not used during inference, we keep it simple and use \ac*{dnn} to predict a factorized Gaussian for the variational posterior distribution on the latent space $\mathbb{R}^{\latentDim}$, \ie $q_{\InferenceParam}(\latentVar|\pcdVar, \graspVar) = \mathcal{N}(\latentVar; \mu_{\InferenceParam}(\pcdVar, \graspVar), diag(\delta_{\InferenceParam}(\pcdVar, \graspVar)))$.

\textbf{Grasp Generation and Shape-Aware Introspection:}
\label{sec:introspection}
For a test object point cloud $\pcdVar^*$, we can generate the corresponding grasps $\graspVar^*$ by performing ancestral sampling: 
\begin{equation}
    \graspVar^* \sim p_{\GeneratorParam}(\graspVar|\latentVar^*) ; \;\; \latentVar^* \sim p_{\GeneratorParam}(\latentVar|\pcdVar^*).
\end{equation}
From Section \ref{sec:intro_nf} we know that we can compute the \textit{exact likelihoods} of $ p_{\GeneratorParam}(\graspVar|\latentVar^*)$ and $p_{\GeneratorParam}(\latentVar|\pcdVar^*)$ from Grasp Flow and Prior Flow, respectively.  
We expect these two quantities to capture the knowledge gained by the model quantitatively. 
With these, our model can be introspective against its unknown knowledge, such as the \textit{incomplete shape} due to partial observation and \textit{unknown object shapes} due to limited coverage of training data.
Specifically, the likelihoods of Grasp Flow quantify the inverse \textbf{view uncertainty} as it sees both the point cloud and the corresponding grasps through the learned latent variable.
Thereby, high values are assigned to grasps towards visible views and low for grasps to the invisible views. 
Complementarily, with the proposed \ac*{dlvm}, Prior Flow is expected to represent the inverse \textbf{object uncertainty} by purely conditioning on the input point cloud, \ie low likelihoods for unknown object shape and high for known ones. 

\vspace{-5pt}
\subsection{Uncertainty-Aware Grasp Evaluation}
\vspace{-5pt}
\label{sec:grasp_evaluator}
To further secure the grasp success, 
we train a discriminative grasp evaluator $f_\psi(\graspVar, \pcdVar)$ that outputs a score to better capture the grasp quality by learning to distinguish feasible grasps from infeasible ones in a supervised manner~\cite{ffhnet}. 
Furthermore, as Grasp Flow is able to quantify the view uncertainty, it is beneficial to incorporate this information into the grasp evaluation. 
With this, we can penalize the grasps approaching the uncertain side of the partially observed object to avoid potential collisions.
To this end, we introduce an uncertainty-aware grasp evaluation strategy that fuses the batch-normalized view uncertainty into the grasp evaluator:
\begin{equation}
 \epsilon f_\psi(\graspVar^*, \pcdVar^*) + (1 - \epsilon) \log{p_{\GeneratorParam}(\graspVar^*|\latentVar^*)},   
\end{equation}
where $\epsilon \in [0,1]$ is a parameter for balancing the effects of grasp quality and reducing potential collisions due to the partial observability (an ablation study in Section 3.2 of the appendix).


\section{Experiment}
\vspace{-5pt}
In this section, we perform comprehensive experiments and analysis to answer the following questions:
\textbf{Q1}: Can the proposed \SecondFlowName facilitate more expressive latent representations for both diverse, high-quality and fast grasp synthesis? \textbf{Q2}: How well does \SecondFlowName perform compared with other state-of-the-art grasp synthesis approaches? 
\textbf{Q3}: How effective is the shape-aware introspection of \SecondFlowName against OOD data and does it benefit grasp evaluation? \textbf{Q4}: Can the model generalize to the complex real-world scenarios (\eg confined and cluttered scenes)? More ablation studies are included in appendix due to space limit.
\vspace{-5pt}
\subsection{Experimental Setup}
\begin{figure}[htbp]
    \centering
    \begin{subfigure}{0.48\textwidth}
        \vspace{-45pt}
        \includegraphics[width=\linewidth, height=4.6cm]{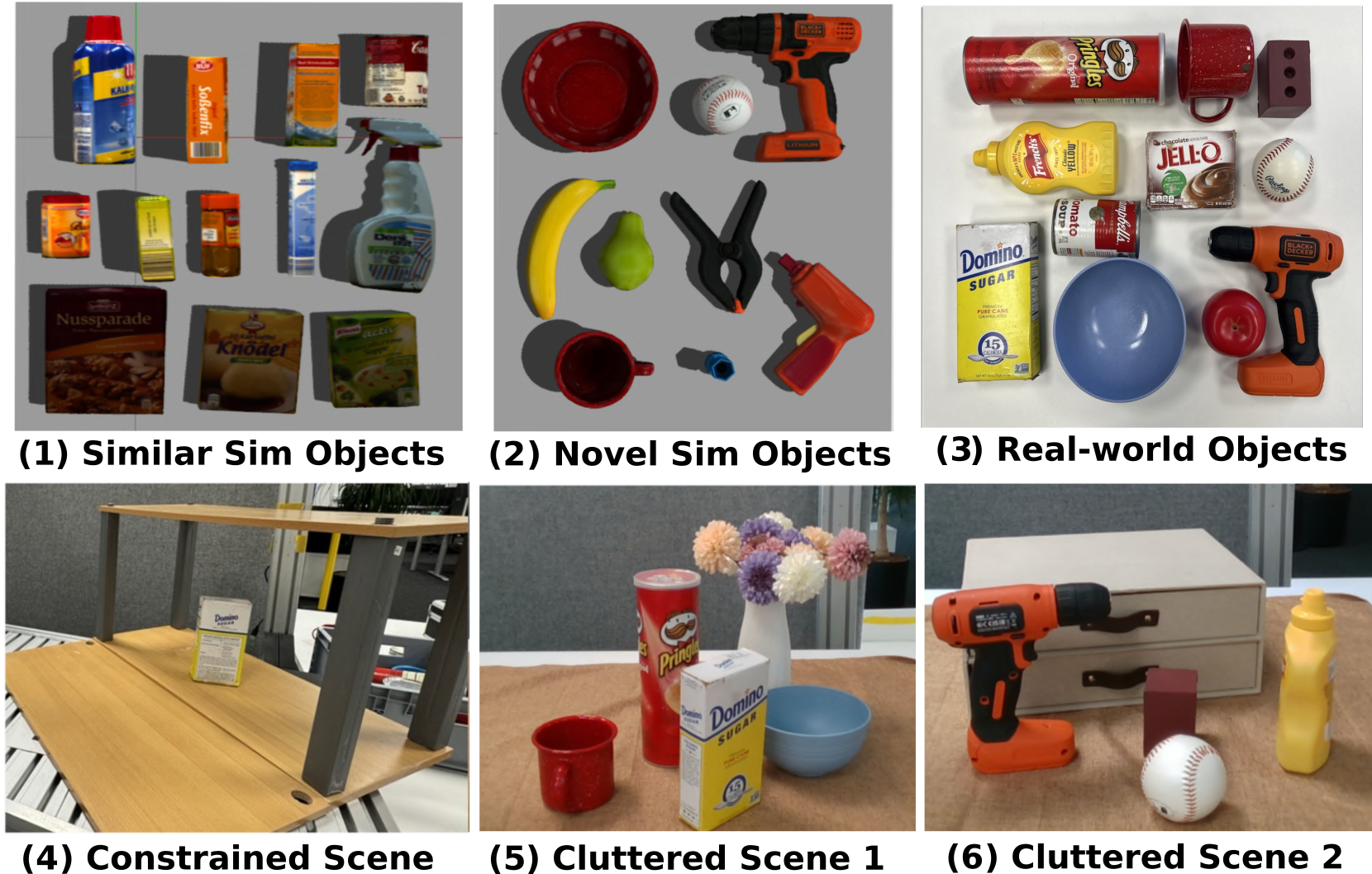}
        \caption{\textbf{Objects and setups evaluated in the simulation and real world.} 
        We test simulation objects with similar (1) and novel (2) shapes, real objects (3), and grasping in confined space (4) and cluttered scene (5-6).}
        \label{fig:YCB_objects}
    \end{subfigure}
    \hfill
    \begin{subfigure}{0.48\textwidth}
        \includegraphics[width=\linewidth, height=4.8cm]{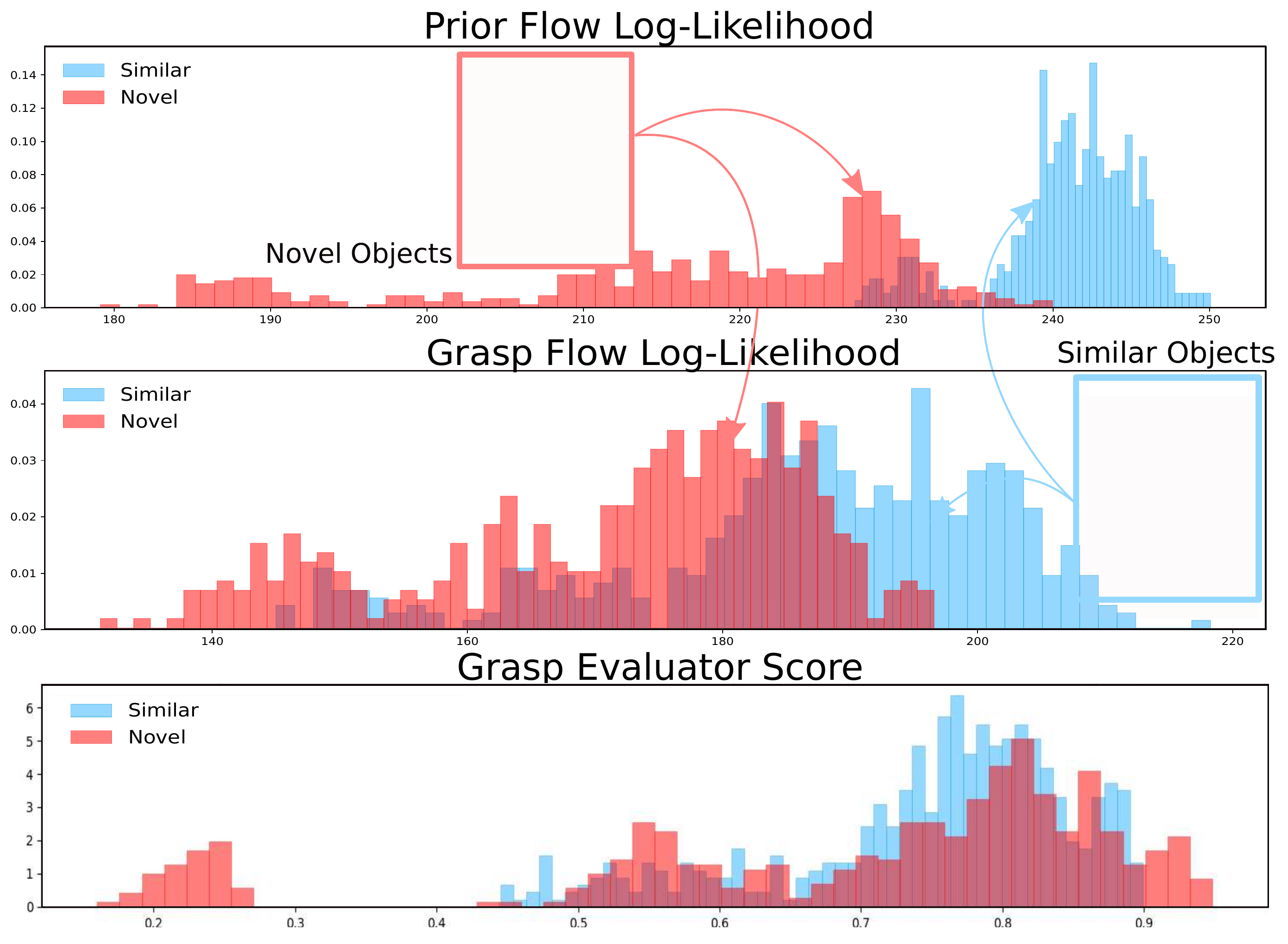}
        \caption{\textbf{Analysis of view and object uncertainty (Top and Middle) and grasp evaluator scores (Bottom)} for In-Distribution (Blue) and Out-of-Distribution Objects (Red). 
        Y axis: frequency/counts, X axis: scores/likelihoods (averaged over predicted grasps for Prior Flow and grasp evaluator). 
        Flow likelihoods can better detect novel objects with low likelihoods.}
        \label{fig:ood_objects}
    \end{subfigure}
    \vspace{-5pt}
    \caption{\textbf{Experiment Setup (a) and Object Uncertainty Evaluation (b).}}
\end{figure}

\vspace{-5pt}
The experiments of grasping single tabletop objects are performed in both the simulation and the real world with the DLR-HIT II hand~\cite{liu2008multisensory}. 
In the real-world, except for single object grasping in unconfined space, we validate our approach in another two complex ones, \ie the confined and cluttered scenarios (\cref{fig:YCB_objects}).
The success criterion is defined as the ability to lift the object 20 cm above its resting position without slippage.
\begin{table} 
\vspace{-0.4cm}
\centering
\caption{Results in Simulation (a) and Ablation Study on Uncertainty-aware Grasping (b).}
    \vspace{-5pt}
\subfloat[Average Success Rate and Run-time in Simulation\label{tab:grasping_simulation_succ}]{
    \ra{1.1}
    \small
    \begin{tabular}{r|rr|r}
    Methods & \makecell{Similar} & \makecell{Novel} &  \makecell{Run time (ms)}  \\ [0.0001cm] 
    \midrule
    Heuristic & $20.9\%$ & $11.1\%$ & $3387$ \\
    \midrule
    cVAE~\cite{ffhnet} & $84.6\%$ & $52.4\%$ & $\textbf{30}$ \\
    GAN~\cite{feng2024dexgangrasp} & $86.0\%$ & $49.4\%$ & $\textbf{30}$ \\
    Diffusion~\cite{weng2024dexdiffuser} & $88.2\%$ & $51.7\%$ & $1610$ \\
    \FirstFlowName & $85.4\%$ & $36.7\%$ & $70$ \\
    
    \SecondFlowName~ & $\textbf{94.6\%}$ & $\textbf{52.7\%}$ & $130$ \\
    \end{tabular}
    }
\subfloat[Results of Uncertainty-aware Grasping\label{tab:ablation_uncertainty_succ}]{
    \small
    \begin{tabular}{r|rr}
    Evaluation Strategy& \makecell{Similar} & \makecell{Novel} \\ [0.0001cm]
    \midrule
    w/o Evaluator & $41.4\%$ & $17.8\%$ \\
    \midrule
    + Evaluator & $90.5\%$ & $50.9\%$ \\
    \makecell{Evaluator +  Prior Flow} & $87.7\%$ & $52.4\%$ \\
     \makecell{Evaluator +  Grasp Flow} & $\textbf{94.6\%}$ & $\textbf{52.7\%}$ \\
    \end{tabular}
    }
\vspace{-20pt}
\end{table}
\textbf{Data:}
We use only simulated data generated based on a heuristic grasp planner for training in Gazebo Simulator~\cite{ffhnet}. 
For \textbf{training}, we use 77 graspable objects filtered from 
KIT~\cite{kasper2012kit} datasets based on their graspability and object type.
For \textbf{testing}, we select objects of two levels of difficulty from the KIT and YCB dataset~\cite{calli2015ycb} in simulation (\cref{fig:YCB_objects}):
\textbf{(1) Similar}: 12 objects from KIT dataset with similar shapes to training objects, serving as \textbf{ID objects}.
\textbf{(2) Novel}: 9 objects from YCB dataset with \textit{shapes distinct from training objects}, often more difficult to grasp, serving as \textbf{OOD objects}.
For real-world evaluation, we use \textcolor{blue}{12} unknown objects from YCB dataset.
(More detail in the appendix.)

\vspace{-5pt}
\subsection{Evaluation in the Simulation}
\vspace{-5pt}
We demonstrate the simulation results through the success rate and runtime in \cref{tab:grasping_simulation_succ}.
\textbf{Baselines:}
1. Heuristic grasp sampler: a heuristic grasp sampler to generate grasps based on the normal of object point clouds~\cite{ffhnet};
2. A \ac*{cvae}-based approach, FFHNet~\cite{ffhnet};
3. A \ac*{gan}-based approach, DexGanGrasp~\cite{feng2024dexgangrasp}; 
4. A \ac*{diffusion}-based approach, DexDiffuser~\cite{weng2024dexdiffuser}.
We apply the same evaluator for all generative grasping baselines except for \SecondFlowName with the proposed uncertainty-aware evaluation strategy.
\textbf{Results:} To answer \textbf{Q1} and \textbf{Q2}, in \cref{tab:grasping_simulation_succ}, first there is a clear performance drop from similar to novel objects, indicating the necessity of having the introspective capability to identify OOD objects.
\SecondFlowName is able to generate high-quality and relatively fast grasp synthesis, outperforming the baselines on both similar objects and novel objects and is 10x faster than the diffusion baseline.
Notably, the diffusion-based approach~\cite{weng2024dexdiffuser} demonstrates relatively better performance compared to cVAE~\cite{ffhnet} on similar objects and GAN~\cite{feng2024dexgangrasp} on both similar and novel objects. 
However, the iterative de-noising process results in a significantly longer runtime ($1610$ms). 

\vspace{-5pt}
\subsection{Uncertainty-Aware Grasp Evaluation}
\vspace{-5pt}
\begin{figure}[t!]
   \centering
   \begin{subfigure}{0.45\textwidth}
       \includegraphics[width=\linewidth, height=3.5cm]{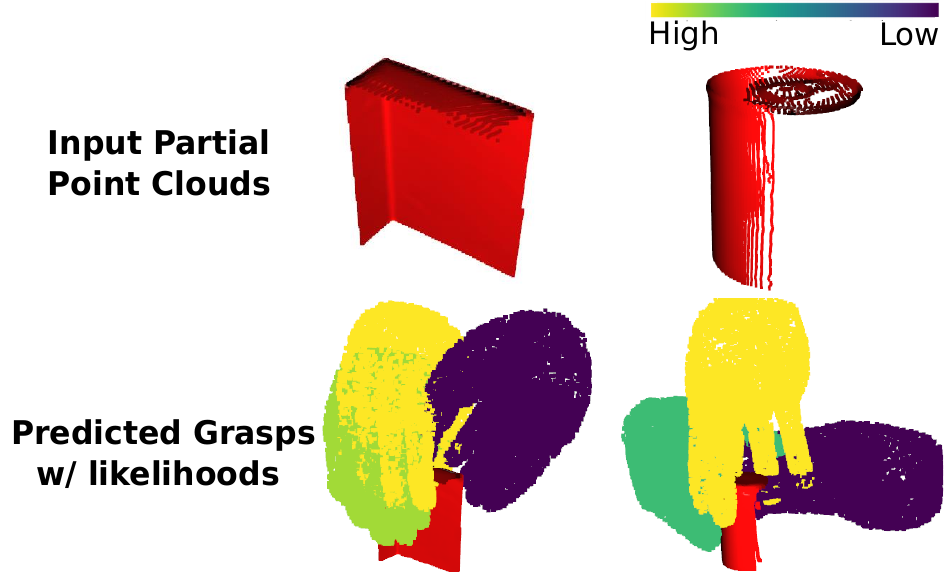}
       \caption{\textbf{Visualizing view uncertainty}. Predicted grasps colored based on Grasp Flow likelihoods: lower for grasps (purple) to the \textit{incomplete} face of the partially perceived point cloud.}
       \label{fig:grasp_flow}
   \end{subfigure}
   \hfill
   \begin{subfigure}{0.52\textwidth}
       \includegraphics[width=\linewidth, height=5cm]{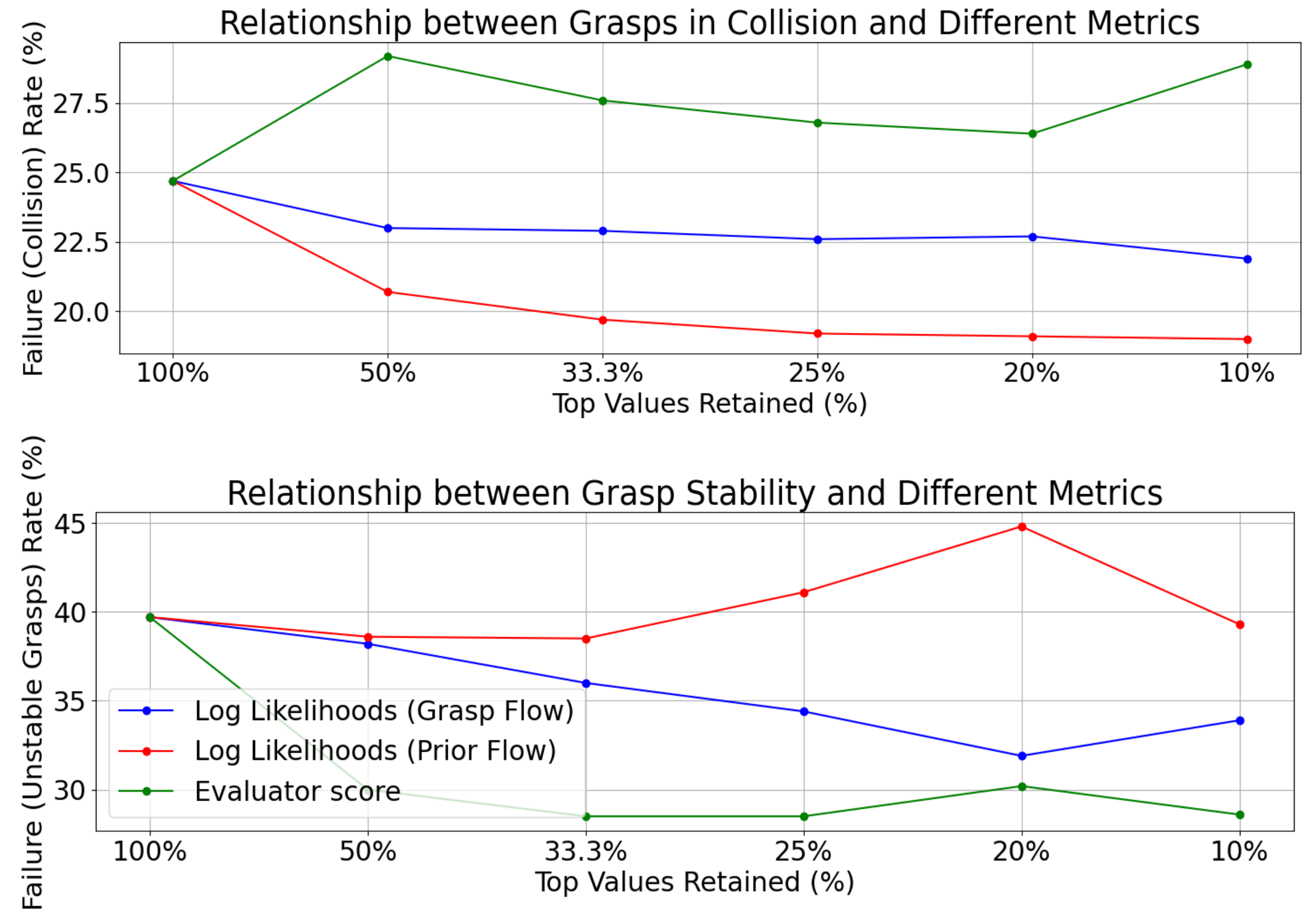}
       \caption{\textbf{Number of collided (Top) and unstable (Bottom) grasps} filtered with an increasing threshold (higher the better). Likelihoods from Grasp Flow (Blue) achieve a more optimal balance between grasp stability and collision.}
       \label{fig:grasp_collision}
   \end{subfigure}
   \vspace{-5pt}
   \caption{\textbf{View Uncertainty Visualization (a) and Evaluation (b).}}
   \vspace{-20pt}
\end{figure}
\paragraph{Uncertainty Quantification:}
To answer \textbf{Q3}, \cref{fig:ood_objects} shows that Prior Flow can better represent the \textbf{object uncertainty} via estimating the density of object shapes. 
It can better distinguish between in-distribution (ID) and novel/OOD objects with distinct shapes to the training data. 
This is useful in situations with multiple objects, where the robot could prioritize grasping in-distribution objects first to \textit{avoid grasp failures}.
On the other hand, Grasp Flow can represent the \textbf{view uncertainty} (visualized in \cref{fig:grasp_flow}). 
To showcase its effects on the predicted grasps, we present the relation between the number of collided and unstable grasps and increasing thresholds of the likelihoods and the grasp evaluator scores (higher the better) in \cref{fig:grasp_collision}. We assess the collision of the generated grasps with
the Flexible Collision Library (FCL) and the grasp stability with Gazebo simulation (more implementation details in the appendix). 
We observe that, in contrast to the grasp evaluator score, both Prior Flow and Grasp Flow demonstrate the ability to reduce collision in the top plot.
In the bottom plot, only the evaluator score and Grasp Flow can help improve grasp stability while the Prior Flow is less effective, which also explains the performance drop in \cref{tab:ablation_uncertainty_succ}.

\textbf{Uncertainty-Aware Grasp Evaluation:}
Grasp Flow likelihoods are utilized for grasp evaluation as it represents the view uncertainty and exhibits a better trade-off between grasp stability and collision avoidance in~\cref{fig:grasp_collision}. 
This capability facilitates grasp selection with higher stability and less collision, leading to greater performance gain (\cref{tab:ablation_uncertainty_succ}). The experiment in~\cref{tab:ablation_uncertainty_succ} follows the same set up as \cref{tab:grasping_simulation_succ} and more details are described in Appendix Section 2.4.
\vspace{-5pt}
\subsection{Evaluation in the Real World}
\vspace{-5pt}

\begin{wraptable}{r}{0.5\textwidth}
\vspace{-1.cm}
\ra{1.4}
\caption{Real-World Experiment Results}
\vspace{-10pt}
\begin{center}
\label{tab:grasping_real_succ} 
\centering
\begin{adjustbox}{width=1.\linewidth}
\begin{tabular}{r|r|r|rr}
Workspace & \multicolumn{1}{c|}{Unconfined} & \multicolumn{1}{c|}{Confined} & \multicolumn{2}{c}{Cluttered} \\
\toprule [0.5pt]
Metrics &\makecell{Avg Succ \\ Rate} & \makecell{Avg Succ \\ Rate} & \makecell{Avg Succ \\ Rate} & \makecell{Clearance \\ Rate} \\ 
\midrule
cVAE~\cite{ffhnet}  & $62.5\%$  & $10.0\%$ & $68.2\%$ & $50.0\%$ \\
\SecondFlowName  & $\textbf{77.5\%}$ & $\textbf{65.0\%}$ & $\textbf{76.0\%}$ & $\textbf{75.0\%}$
\end{tabular}
\end{adjustbox}
\end{center}
\vspace{-20pt}
\end{wraptable}


\textbf{Unconfined Space:} This setup is similar to the simulation, where 8 objects are grasped 10 times each. For \textbf{Q4}, \cref{tab:grasping_real_succ} shows that \SecondFlowName~is able to generalize to the real robot with a smaller gap.
\textbf{Confined Space:} We selected a confined space, \ie a two-tier shelf to further mimic the realistic scenarios (d) in \cref{fig:YCB_objects}.
In \cref{tab:grasping_real_succ}, a large performance gain is achieved by \SecondFlowName when compared to cVAE. 
The main failure of cVAE is due to its biased top grasps colliding with the environment (shelf). 
In such a confined space, a \textit{diverse} grasp synthesizer is more effective than the mode-seeking cVAE.
\textbf{Cluttered Scenes:}
We conduct grasping experiments in cluttered scenes 
as (e) and (f) shown in~\cref{fig:YCB_objects}. The metrics Clearance Rate (CR) denotes the probability of robots clear the scene. The predicted grasps with collision are filtered out based on FCL. 
\SecondFlowName achieves a higher success rate (SR) and also clearance rate (CR). More details in the appendix.

\vspace{-5pt}
\section{Conclusion}
\vspace{-5pt}
We introduce a novel flow-based variational approach, \SecondFlowName for generative grasp synthesis with better quality and diversity. 
This is achieved by mitigating the insufficiently informative latent features when applying \ac*{cnf} directly and overcoming problems in \ac*{cvae}-based approaches, \ie, mode-collapse and the mis-specified prior, as well as inefficiency issues from diffusion-based approaches.
Moreover, the model is equipped with shape-aware introspection quantified by the exact flow likelihoods, which further facilitates a novel hybrid scheme for uncertainty-aware grasp evaluation.
Comprehensive experiments in the simulation and real world demonstrate strong performance and efficiency.

\section{Limitations}
\vspace{-5pt}
Though our proposed idea exhibits numerous strengths, its limitations remain to solve for further improvement in the future. We highlight the most pronounced limitations to facilitate future research.
\textbf{(1)} Trade-off between run-time and performance: 
reducing the flow size can improve the run-time but at expense of a slight performance loss. Investigating how to strike a balance is relevant for broader robotic applications.
\textbf{(2)} The lack of abilities for efficient adaptation towards objects that differ significantly from those in the training dataset, which is hard to avoid for robots deployed in the wild~\cite{feng2019,feng2022, noseworthy_shaw_icra24}. We can further scale up the synthetic dataset, leveraging GPU-powered simulation such as Isaac Gym/Sim to increase generalization capability.
\textbf{(3)} Sim-to-Real gap: Though the objects in simulation and real world are different in evaluation, a success rate drop of $17.1\%$ encourages us to further investigate how to eliminate the sim-to-real gap, through various approaches such as camera noise modeling, physical parameters randomization and better simulators.




\bibliography{ref}  
\includepdf[pages=-]{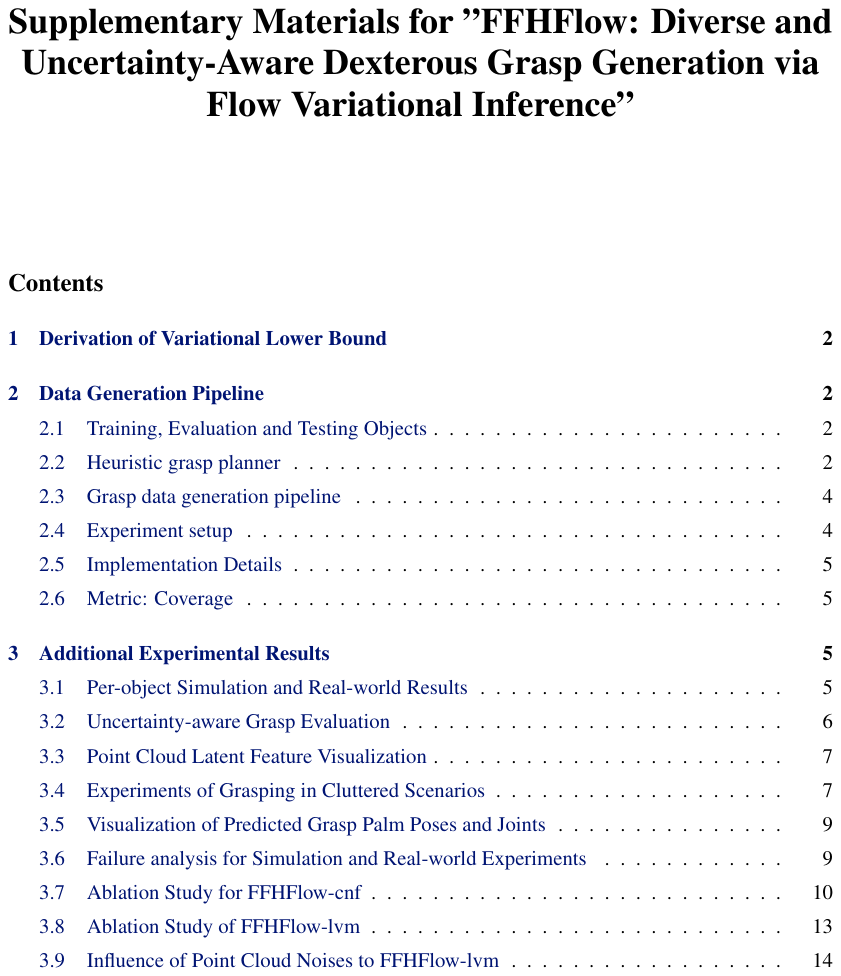}
\end{document}